**LUNAR SURFACE IMAGE RESTORATION USING U-NET BASED DEEP NEURAL NETWORKS.**
H. Roy[1,4], S. Chaudhury[2], T. Yamasaki[2], D.M. DeLatte[3], M. Ohtake[4], T. Hashimoto[1,4] [1]The University of Tokyo, Department of Electrical Engineering and Information Systems, [2]The University of Tokyo, Department of Information and Communication Engineering, [3]The University of Tokyo, Department of Aeronautics and Astronautics, 7-3-1 Hongo, Bunkyo, Tokyo 113-8654, [4]Institute of Space and Astronautical Science / Japan Aerospace Exploration Agency, (ISAS/JAXA), 3-1-1 Yoshinodai, Chuo-ku, Sagamihara City, Kanagawa, 252-5210, Japan. hiya.roy@ac.jaxa.jp

**Introduction:** In the quest of exploring and understanding the planetary body, Moon, several missions such as Clementine, Japanese SELenological and ENgineering Explorer (SELENE/KAGUYA), NASA's Lunar Reconnaissance Orbiter (LRO) and so on, have been carried out over the years. The major goals of these missions are to gather knowledge about different aspects such as the origin and evolution of Moon, the mineral composition of the lunar surface etc. With the advancement in imaging technology, different types of onboard imaging cameras of all of these missions can provide high resolution lunar surface images that can yield additional information about the lunar surface. However, in order to obtain high resolution images, the swath width of the satellite is kept lower which in turn creates discontinuity in the lunar surface image creating some missing pixel region. Our aim is to intelligently predict the lunar surface image in a context-aware fashion to fill in the missing pixels so as to obtain the entire surface image. Here, we use the grayscale image of the lunar surface captured by Multiband Imager (MI) onboard Kaguya satellite. However this lunar surface image also has some vertical blacked stripped errors (amounting to less than 2% of the entire captured image) and we term them as missing pixels as shown in Figure 1.

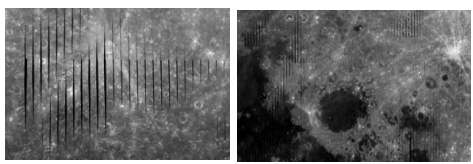

Fig 1: Zoomed in and zoomed out version of lunar surface image with black stripped errors

**Reason for the missing pixels:** Multiband Imager (MI) instrument onboard the Kaguya satellite continuously observe the lunar surface. The swaths of visible bands are designed to be 19.3 km from SELENE's 100 km altitude [1]. The satellite and the moon are rotating around their own axes at the same time. Therefore, it is found that, by the time the satellite finishes one complete rotation around its own orbit, the moon already moves away 33.4 km from its previous position. This makes it difficult for the instrument to capture the entire surface image, thus creating missing pixels on the captured image as shown in Figure 1.

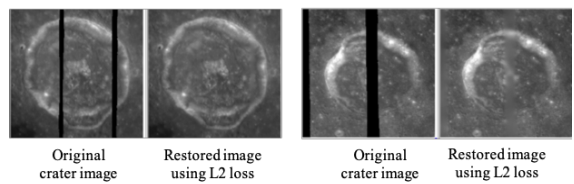

Fig 2: Crater image restoration

**Motivation:** The main motivation behind restoring the lunar surface image by filling in the missing pixels (as shown in Figure 2) is to analyze the morphological features such as the size of the crater, the size of the central peaks, their interior structure etc. on the lunar surface and compare these with other planetary bodies.

**Data:** In this work, we have used the averaged lower resolution mosaic data of the lunar surface captured by Multiband Imager onboard the JAXA lunar explorer satellite SELENE (Kaguya). The image covered the whole lunar surface with the longitude ranging from +180 degree to -180 degree and latitude ranging from +85 degree to -85 degree. The image consists of 46080x21760 pixels with an image size of 600MB. We converted the longitude and the latitude of the crater locations in terms of pixel values, where each degree is considered to be 128 pixels and cropped the crater images with and without the black stripes. We have 340 crater images with no missing pixels (which we term as 'clean crater' henceforth) and 53 images of craters with black strips or missing pixels (which we term as 'corrupted crater' hereafter). After that we resized each crater images to 256x256 pixels and created the pairs of clean crater image and the corrupted crater images by randomly superimposing the black strips on the clean crater images. We used 10000 image pairs for training the network, and 5000 image pairs for validation and testing purposes.

**Proposed concept:** In this work, our goal is to learn the mapping between the clean and corrupted crater image and generate clean image for an unknown corrupted crater image. Since deep convolutional neural network architectures have proved to be successful for image restoration tasks [2,3] in recent years, we adopt the neural network architecture known as U-Net proposed by Ronneberger et al. [4] for the lunar surface restoration. This network has the capability of generating an output image from an input image by learning the mapping in an end-to-end fashion. The schematic of the proposed concept is shown in Figure 3, where we extract the black strips randomly from a crater image having the black strips and superimpose it on a clean crater image to artificially generate a pair of clean and corrupted crater image and then train the adopted U-net in a supervised way to learn the mapping between these clean and corrupted pairs. Later we validate and test the network with an unknown corrupted crater image to generate its corresponding clean image to finally restore the lunar surface image.

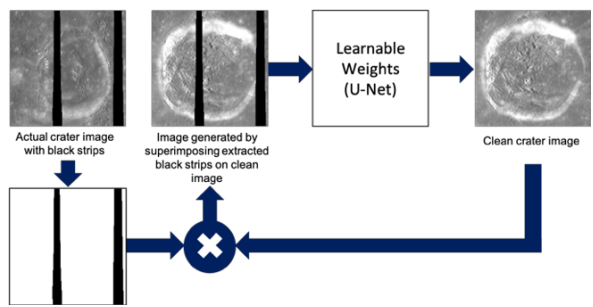

Fig 3: Schematic diagram of the proposed concept

**Training details:** The adopted U-Net architecture is a combination of a contraction and an expansion path where the contraction path consist of a sequence of convolution operations followed by a non-linear activation and a max-pooling operation and the expansion path consist of a sequence of up-convolutions and concatenations with high resolution features from the contracting path [4]. For learning the mapping between the clean and corrupted crater image, we adopt the training methodology of minimizing standard L2 loss. It took around 12 hrs to train the network on a TITAN-V GPU. We implemented our model using Pytorch in Python.

**Experimental results:** To verify the effectiveness of our proposed algorithm we perform both qualitative and quantitative evaluation. We calculate the PSNR values of the restored images compared to their original uncorrupted images. Figure 4 shows the experimental results where the first column depicts the artificially generated corrupted images along with their PSNR values, the second column shows the restored images with their corresponding PSNR values and the third column represents the original crater image. Here we can see that the restored images have better PSNR values compared to their corresponding corrupted images. Qualitative analysis from figure 4 shows that the predicted pixel values seem natural to the surrounding pixel values.

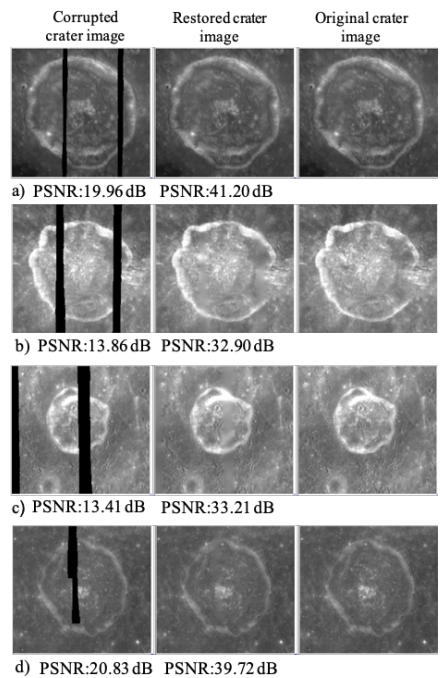

Fig 4: Quantitative results of restored crater images

**Conclusion:** In this paper we show that the missing pixels of the lunar surface image can be restored successfully using U-Net based deep neural networks. Experimental evaluations show that the proposed deep learning approach can restore missing pixels with impressive quality for the grayscale images. To the best of our knowledge the application of image restoration technique on planetary surface image is a new approach to address the concerns of the planetary scientists for precise analysis of planetary surfaces. In future, we plan to extend this work by incorporating additional information using multiband images.